\documentclass{article}
\pdfoutput=1
\PassOptionsToPackage{numbers, compress}{natbib}

\usepackage[preprint]{dub3d}

\usepackage[utf8]{inputenc} 
\usepackage[T1]{fontenc}    
\usepackage{hyperref}       
\usepackage{url}            
\usepackage{booktabs}       
\usepackage{amsfonts}       
\usepackage{nicefrac}       
\usepackage{microtype}      
\usepackage{xcolor}         

\usepackage{multirow}
\usepackage{multicol}
\usepackage{graphicx}
\usepackage{subfigure}
\usepackage{caption}
\usepackage{array}

\usepackage{wrapfig}
\usepackage{tikz}

\title{Distinguish Any Fake Videos: Unleashing the Power of Large-scale Data and Motion Features}

\author{
Lichuan Ji\textsuperscript{1} \quad Yingqi Lin\textsuperscript{1} \quad
Zhenhua Huang\textsuperscript{1} \quad Yan Han\textsuperscript{1} 
\quad Xiaogang Xu$^{2,3}$$^*$ \vspace{0.1in}\\ \textbf{Jiafei Wu}$^{1}$$^*$ \quad \textbf{Chong Wang}$^4$ \quad \textbf{Zhe Liu}\textsuperscript{1}\vspace{0.1in}\\
$^1$\textit{Zhejiang Lab}\quad 
 $^2$\textit{The Chinese University of Hong Kong}  \quad $^3$\textit{Zhejiang Univeristy} \quad  $^4$\textit{Ningbo Univeristy} \vspace{0.1in}\\ 
{\tt \small \{jilichuan, linyq, huangzhenhua, hanyan, wujiafei, zhe.liu\}@zhejianglab.com}\vspace{0.1in}\\ {\tt \small xiaogangxu00@gmail.com} \quad {\tt \small wangchong@nbu.edu.cn} \vspace{0.1in}\\
$*$ indicates the corresponding author}

\begin{document}

\maketitle
\begin{abstract}

The development of AI-Generated Content (AIGC) has empowered the creation of remarkably realistic AI-generated videos, such as those involving Sora. However, the widespread adoption of these models raises concerns regarding potential misuse, including face video scams and copyright disputes. Addressing these concerns requires the development of robust tools capable of accurately determining video authenticity. The main challenges lie in the dataset and neural classifier for training. 
Current datasets lack a varied and comprehensive repository of real and generated content for effective discrimination. 
In this paper, we first introduce an extensive video dataset designed specifically for AI-Generated Video Detection (GenVidDet). It includes over \textbf{2.66 M} instances of both real and generated videos, varying in categories, frames per second, resolutions, and lengths. The comprehensiveness of GenVidDet enables the training of a generalizable video detector. We also present the Dual-Branch 3D Transformer (DuB3D), an innovative and effective method for distinguishing between real and generated videos, enhanced by incorporating motion information alongside visual appearance. DuB3D utilizes a dual-branch architecture that adaptively leverages and fuses raw spatio-temporal data and optical flow. We systematically explore the critical factors affecting detection performance, achieving the optimal configuration for DuB3D. Trained on GenVidDet, DuB3D can distinguish between real and generated video content with \textbf{96.77\%} accuracy, and strong generalization capability even for unseen types.

\end{abstract}
\section{Introduction}

The evolution of AI-Generated Content (AIGC) techniques has revolutionized content creation, enabling users to generate videos from simple text prompts~\cite{videoworldsimulators2024,pikaPika}. This ease of video generation has led to a significant increase in AI-generated videos, enriching the diversity of media content. However, the widespread use of these models has raised concerns about potential misuse, such as face video scams~\cite{ferrara2024genai} and copyright disputes~\cite{harvod}. To address these issues, robust tools are needed to accurately verify the authenticity of video content. This task, known as ``Distinguish Any Fake Videos" (DAFV), is a critical challenge in the AIGC field.

To implement DAFV, two main challenges arise concerning the dataset and classifier for training. First, current datasets lack a diverse and extensive collection of both real and generated content.
They are either dominated by real videos~\cite{wang2024internvid} or synthetic videos~\cite{wang2024vidprom} with confined types.
This limitation hampers the ability to train robust classifiers that can generalize well to unseen data. Additionally, existing video detectors~\cite{liu2021video,fan2020pyslowfast} are designed for general video tasks and may omit crucial feature modeling mechanisms. There is a lack of specialized neural network architectures tailored to distinguish between features encompassing both real and generated content.

\begin{figure}
  \centering
  \includegraphics[scale=0.5]{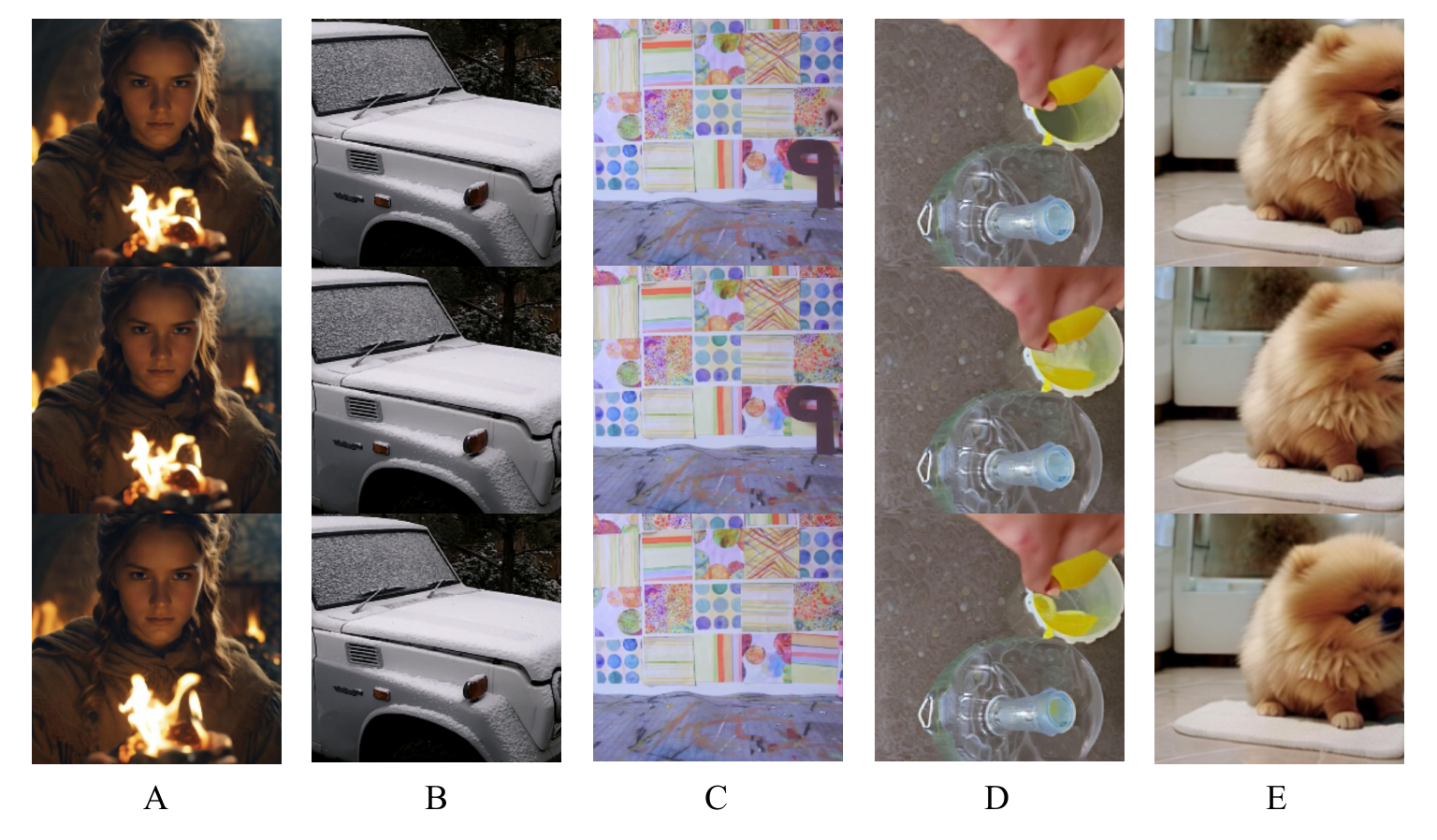}
  \vspace{-0.15in}
  \caption{Can you distinguish which are real videos? For every choice, we extract 3 frames (interval with 4 frames) from real and generated videos, and the answer is in the Appendex \ref{sec:appendex_game}.}
  \label{fig:game}
  \vspace{-0.4in}
\end{figure}

In this paper, we introduce GenVidDet, a meticulously curated video dataset designed specifically for detecting AI-generated video content. It offers a diverse array of real and generated videos to reflect the complexity of modern video content. Additionally, we present the Dual-Branch 3D Transformer (DuB3D), an innovative method for distinguishing between real and generated videos. DuB3D represents a paradigm shift in fake video detection by integrating motion information with visual appearance, providing a more nuanced and comprehensive analysis of video content.

Regarding the dataset, we focused on two key aspects that differ significantly from existing video datasets. First, we considered critical variables in video types beyond category, such as video resolution and length.
For real videos, we curated them from various existing datasets, carefully filtering out low-quality samples. We then supplemented these with AI-generated videos to enhance the diversity in categories, scenarios, frame rates, resolutions, and lengths. Specifically, we used the latest video generators to produce additional samples.
Our dataset includes over 2.66 million instances of both real and generated videos, achieving a balanced ratio of positive and negative samples. Empirical evidence shows that the comprehensiveness of GenVidDet facilitates the training of a highly generalizable video detector.

In terms of network architecture, we recognized the pivotal role of motion information in enhancing recognition accuracy, especially for unseen samples. Existing video generators have advanced appearance modeling, making identification with raw video content alone challenging. However, motion modeling in current video generators is either sub-optimal or inconsistent with appearance modeling mechanisms. As a result, the motion patterns in real and fake videos differ more distinctly, providing easier characteristics to distinguish.
Accordingly, we designed a dual-branch architecture, DuB3D, based on the 3D transformer framework~\cite{liu2021video}. This innovative network leverages the 3D transformer's capability to extract information from both raw spatio-temporal content and optical flow, integrating visual appearance and motion data harmoniously. By adopting a dual-pathway design, DuB3D effectively fuses spatio-temporal data and optical flow, enhancing recognition performance.

Additionally, we conducted a systematic analysis of the critical factors influencing detection performance to achieve the optimal configuration for DuB3D. By examining the impact of input frame rate and the correlation between motion and frame matching, we aim to assist the academic community in better understanding and addressing issues in DAFV.

Extensive experiments are conducted on our GenVidDet datasets with DuB3D architecture. DuB3D can distinguish between real and generated video content with \textbf{96.77\%} accuracy, and strong generalization capability even for unseen types. In summary, our contribution is three-fold.
\begin{itemize}
    \item[$\bullet$] We construct a new comprehensive dataset for fake video detection.
    \item[$\bullet$] We highlight the importance of motion cues in our study and have crafted a dual-branch network architecture to harness these cues effectively.
    \item[$\bullet$] We conduct extensive analyses to validate the generalizability of our network and identified key factors that significantly influence this generalizability.
\end{itemize}
\label{sec:intro}

\vspace{-0.1in}
\section{Related Work}
\label{sec: Related Work}
\vspace{-0.1in}

\subsection{Existing Large-scale Video Datasets}

\noindent\textbf{Generated Video Datasets.}
Currently, publicly available datasets focusing on text-to-video prompts are rare. For example, VidProM~\cite{wang2024vidprom} aims to build the first text-to-video prompt-gallery dataset, comprising 1.67 million unique text-to-video prompts and 6.69 million videos generated by four state-of-the-art diffusion models. However, no real videos are involved in VidProM. 
This paper utilizes VidProM as one of the generated video datasets for constructing our GenVidDet dataset. Unlike current datasets, GenVidDet includes multiple generated data sources with variations in categories, frame rates, resolutions, and lengths. Consequently, GenVidDet enables the training of a highly generalizable video detector. We hope the collection of diffusion-generated videos will be valuable for further research in AI-generated video detection.

\noindent\textbf{Real Video Datasets.}
Many real video datasets have been proposed. For example, InternVid~\cite{wang2024internvid} is a large-scale, video-centric multimodal dataset containing over 7 million videos totaling nearly 760,000 hours.
HD-VG-130M~\cite{videofactory} comprises 130 million high-definition, widescreen, watermark-free text-video pairs from the open domain. HowTo100M~\cite{miech2019howto100m} is a large-scale dataset of 136 million narrated video clips with captions, focusing on instructional videos from YouTube, covering 23,000 activities across various domains.
HD-VILA100M~\cite{xue2022hdvila} is a large-scale, high-resolution video-language dataset designed for multimodal representation learning, containing 3.3 million high-quality videos distributed evenly across 15 categories. WebVid-10M~\cite{Bain2021FrozenIT} addresses the low video-text correlation issue with 10 million high-quality video-text pairs collected from stock footage websites. Our GenVidDet dataset amalgamates various real video datasets, preprocessing them to harmonize input requirements.

\vspace{-0.1in}
\subsection{Video Diffusion Models}

\label{sec: Video Diffusion Models}
Significant advancements have emerged in video diffusion models, with notable methods in both Text-to-Video and Image-to-Video synthesis. Text-to-video synthesis, a relatively new research direction, has recently demonstrated remarkable generative power for producing high-quality video content from textual prompts. For example, Pika~\cite{pikaPika}, a commercial text-to-video model by Pika Labs, has a significant influence in the field. Text2Video-Zero~\cite{khachatryan2023text2videozero} employs zero-shot video generation with textual prompts. Sora~\cite{liu2024sora}, released by OpenAI, can produce up to 1-minute high-quality videos adhering closely to user instructions. VideoCrafter2~\cite{chen2024videocrafter2} trains high-quality video diffusion models without using high-quality videos. ModelScope~\cite{wang2023modelscope} evolves from a text-to-image model by incorporating spatio-temporal blocks for consistent frame generation and smooth movement transitions.
In addition to Text-to-Video models, which rely solely on text prompts, there are emerging Image-to-Video models. For instance, DynamiCrafter~\cite{xing2023dynamicrafter} uses visual content to animate open-domain images, and I2VGen-XL~\cite{zhang2023i2vgenxl} generates high-quality videos from a single static image. 

\vspace{-0.1in}
\subsection{Video Detection Model}

Video detection models have become powerful tools for distinguishing between generated and real content. Several notable models in current research include DeepFake~\cite{altuncu2022deepfake}, which demonstrated the effectiveness of mixed convolutional transformer networks for detection tasks. Video Swin-Transformer~\cite{liu2021video} employs a pure-transformer architecture for video recognition, leveraging spatiotemporal locality inductive bias. Timesformer~\cite{fan2020pyslowfast} adapts the standard Transformer architecture to video by enabling spatiotemporal feature learning directly from a sequence of frame-level patches.

\section{Data Curation}

This section begins with a comprehensive exposition of the composition of the GenVidDet dataset, encompassing both real and AI-generated videos (Section~\ref{Sec:Composition of the GenVidDet}). Subsequently, we elucidate the procedural steps employed during the curation of the GenVidDet dataset, underscoring two critical phases: Video Collection (Section~\ref{Sec:Collecting-Videos}) and Video Generation (Section~\ref{Sec:Generating Videos}). Lastly, we furnish a meticulous analysis of the GenVidDet dataset, offering insights into its statistical attributes and outlining its potential applications for future research (Section ~\ref{sec:Statistics and Analysis}).

\vspace{-0.1in}
\subsection{Overview for the Composition of GenVidDet}
\label{Sec:Composition of the GenVidDet}
\vspace{-0.1in}

We have constructed the GenVidDet dataset, which is a large-scale videos dataset that encompasses both real-world videos and AI-generated videos produced by eight distinct video generation models, as illustrated in Table \ref{tab:GenVidDet-Dataset}. The dataset consists of more than 2.66 million video clips, accumulating a total runtime of over 4442 hours. The extensive and diverse collection establishes a comprehensive dataset for detecting AI-generated videos. 

The construction of the GenVidDet dataset involved two methods: collecting videos from existing datasets and generating videos autonomously using open video generation models. The subsequent sections will provide further details on these approaches.

\begin{table}
  \caption{Composition of the GenVidDet Dataset}
  \label{tab:GenVidDet-Dataset}
  \centering
  \resizebox{1.0\linewidth}{!}{
  \begin{tabular}{p{3.0cm}<{\centering}p{2.4cm}<{\centering}p{3.2cm}<{\centering}p{3.2cm}<{\centering}p{2.4cm}<{\centering}}
    \toprule
    Category & Method & Source & Model & Count \\
    \midrule
    \multirow{2}{*}{Real Videos} & Collection   & InternVid & - & 1,178,838 \\
    & Collection & HD-VG-130M    &   -  & 286,125 \\
    \midrule
    \multirow{8}{*}{AI-Generated Videos}  & Collection & VidProM & Pika & 287,997 \\
    & Collection & VidProM & ModelScope & 262,787\\
    & Collection & VidProM & Text2Video-Zero & 288,314 \\
    & Collection & VidProM & VideoCraft2 & 286,384 \\
    & Generated & -  & Open-Sora  & 23,761 \\
    & Generated & -  & Open-Sora-Plan  & 8,387 \\
    & Generated & -  & DynamiCrafter & 39,999 \\
    & Generated & -  & StreamingT2V & 720 \\
    \bottomrule
  \end{tabular}}
  \vspace{-0.2in}
\end{table}
\vspace{-0.1in}

\vspace{-0.1in}
\subsection{Collecting Videos: Carefully Screened Ensemble of Current Datasets}
\label{Sec:Collecting-Videos}
\vspace{-0.1in}

The dataset collection process is divided into two primary components: the collection of real videos and the collection of AI-generated videos.
Our principle during collection is acquiring diverse videos with high quality and filter out the bad ones.

\noindent\textbf{Collecting Real Videos.}
In the process of collecting the real video dataset, we initiated our collection from the InternVid~\cite{wang2024internvid} dataset. Specifically, we targeted InternVid-10M-FLT which carefully curated for quality, contains 10 million videos resized to 256 pixels (short edge). From this dataset we selected 1 million videos firstly. To enhance the resolution of our dataset, we also handpicked nearly 180,000 clips from Internet by URLs in this dataset. Additionally, we incorporated the HD-VG-130M~\cite{videofactory} dataset, a extensive collection from the open domain featuring 130 million high-definition, widescreen, watermark-free text-video pairs. To prevent distribution bias in our dataset, we randomly sampled 286,000 high-resolution videos from HD-VG-130M, further enhancing the authenticity and diversity of our collection.

\noindent\textbf{Collecting AI-Generated Videos.}
The AI-generated video dataset was collected from the VidProM~\cite{wang2024vidprom}  dataset, encompassing videos generated by four models: Pika~\cite{pikaPika}, ModelScope~\cite{wang2023modelscope}, Text2Video-Zero~\cite{khachatryan2023text2videozero}, and VideoCraft2~\cite{chen2024videocrafter2}. The dataset comprises approximately 1.12 million video clips, with a roughly equal distribution across the models, ensuring diversity while maintaining balance among the different generative models. This approach not only expands the coverage of generative models but also guarantees a wide representation of types and styles in the training dataset. It provides a solid foundation for training more accurate and robust video detection models.

\subsection{Generating Videos: More Newest Video Generators}
\label{Sec:Generating Videos}
\vspace{-0.1in}

\begin{table}
  \caption{Self-generated Videos Curation}
  \label{tab:Generating-Videos-Curation}
  \centering
    \resizebox{1.0\linewidth}{!}{
  \begin{tabular}{p{3cm}<{\centering}p{3cm}<{\centering}p{2.5cm}<{\centering}p{2.5cm}<{\centering}p{2.5cm}<{\centering}}
    \toprule
    Method & Model & Resolution & Frames Rate& Frames \\
    \midrule
    \multirow{3}{*}{Text-to-Video} & Open-Sora   & 256P  & 8fps & 16 \\
    & Open-Sora & 512P  & 8fps & 16  \\
    & Open-Sora-plan & 256P & 24fps& 65 \\
    & StreamingT2V & 720P & 10fps & 24 \\
    \midrule
    \multirow{1}{*}{Image-to-Video}  & DynamiCrafter & 256P & 8fps & 16 \\
    \bottomrule
  \end{tabular}
  }
  \vspace{-0.2in}
\end{table}

Although the VidProM dataset contains a substantial number of AI-generated video samples, it is restricted to the four video generation models mentioned earlier, lacking some important generator's samples, e.g., the latest Sora~\cite{liu2024sora} model.
Hence, to enhance the GenVidDet dataset, we autonomously generated videos using the newest video generation models. 

Table \ref{tab:Generating-Videos-Curation} illustrates the details of our employed generators. We employed three Text-to-Video models and one Image-to-Video model. Specifically, we employed the Open-Sora~\cite{hpcaitechOpenSora} model to generate videos at two different resolutions, 256P and 512P. The Open-Sora-Plan~\cite{pkuyuanlabOpenSoraPlan} model was utilized to generate high frame rate videos at 24 fps. Furthermore, we enhanced the generation of 720P high-resolution videos by combining the ModelscopeT2V~\cite{wang2023modelscope} as the base model with the StreamingT2V~\cite{henschel2024streamingt2v} model. Additionally, we utilized the DynamiCrafter~\cite{xing2023dynamicrafter} model to perform image-to-video generation. 

\begin{wrapfigure}{r}{0.4\textwidth}
  \centering
  \vspace{-0.2in}
  \includegraphics[width=1\linewidth]{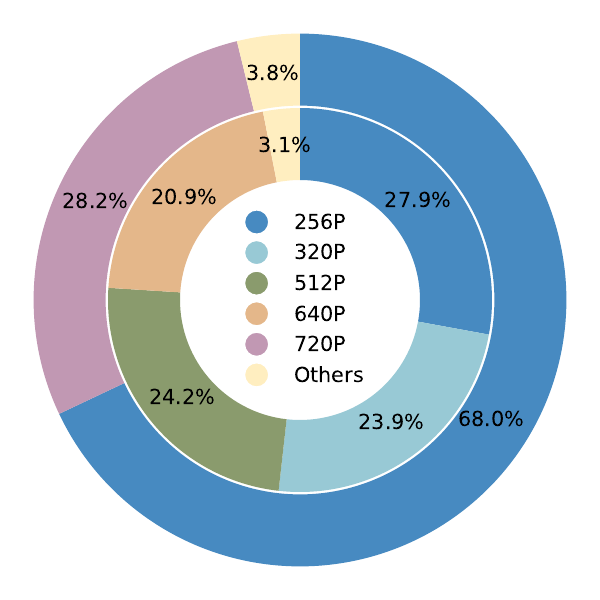}
  \vspace{-0.2in}
  \caption{The resolution distribution differences. Outer ring for real videos, while inner ring for generated videos.}
  \label{fig:videos-resolution}
  \vspace{-0.3in}
\end{wrapfigure}

We scrupulously designed this combination to encompass various resolutions, frame rates, and generation methods, thereby effectively demonstrating the capabilities of latest video generation models. While the number of self-generated videos is currently limited, \textit{they are specifically designed to evaluate model performance on a wider range of video content beyond the training data}. 
This generalization evaluation set ensures that the models demonstrate proficiency not only within the training dataset but also across different video materials.

\vspace{-0.1in}
\subsection{Statistics and Analysis}
\label{sec:Statistics and Analysis}
\vspace{-0.1in}

\begin{wrapfigure}{r}{0.4\textwidth}
  \centering
  \vspace{-0.2in}
  \includegraphics[width=1\linewidth]{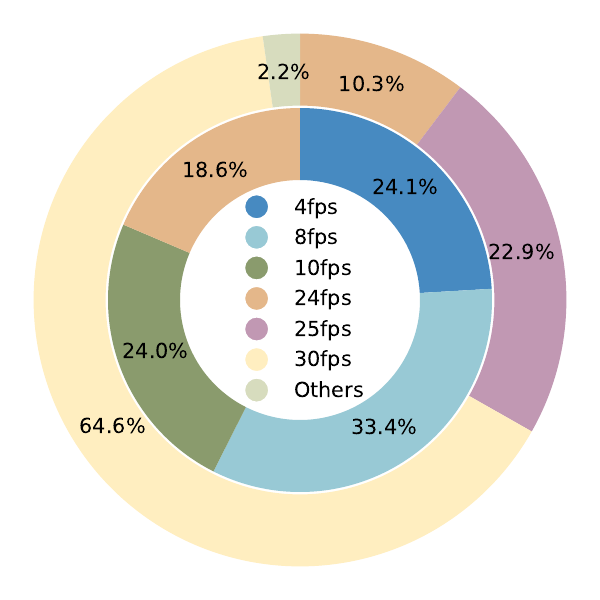}
  \vspace{-0.2in}
  \caption{The frame rate distribution differences. Outer ring for real videos, while inner ring for generated videos.}
  \label{fig:videos-fps}
  \vspace{-0.2in}
\end{wrapfigure}

To acquire a comprehensive understanding of GenVidDet's composition, we have conducted a meticulous analysis of both real and generated videos across various parameters, including resolution, frame rate, and frame count.

\noindent\textbf{Resolution Analysis.}
Figure \ref{fig:videos-resolution} displays the resolution distribution within our dataset. Remarkably, The real video dataset primarily consists of 256P and 720P videos. The 256P videos are sourced from InternVid~\cite{wang2024internvid}, while the 720P and other resolutions are collected from Internet by URLs in InternVid~\cite{wang2024internvid} and HD-VG-130M~\cite{videofactory}. In contrast, the videos that generated by a diverse array of video generation models exhibit a more diverse composition of resolution, characterized by a more evenly spread distribution.

\noindent\textbf{Frame Rate Analysis.}
Figure \ref{fig:videos-fps} provides a more detailed breakdown of the frame rate statistics. In the case of real videos, represented by the outer ring, those operating at a rate of 30 fps are predominant, with a significant portion, approximately 64.6\%. In addition, a combined 33.2\% of videos fall within the medium frame rate categories of 24 fps and 25 fps, suggesting a somewhat concentrated distribution. On the other hand, the generated videos, as illustrated by the inner ring, display a more evenly distributed frame rate profile. The frame rates of 4 fps, 8 fps, 10 fps, and 24 fps each occupy roughly equivalent proportions, indicating a more balanced distribution.

\begin{figure}[t]
  \centering
  \vspace{-0.2in}
  \begin{minipage}{0.48\textwidth}
    \includegraphics[width=\textwidth]{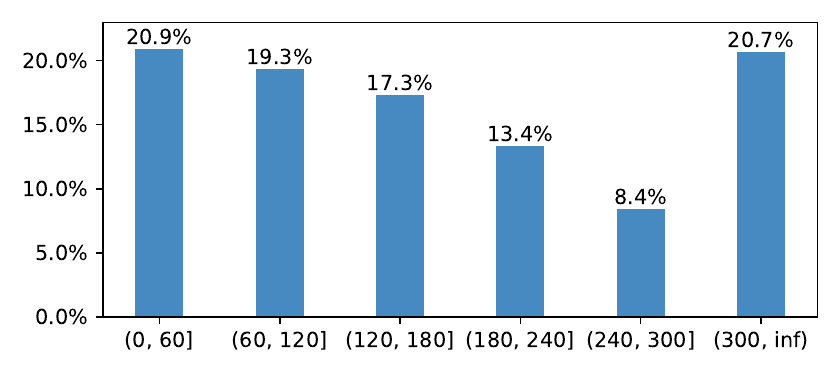}
    \vspace{-0.2in}
    \caption*{(a) real videos}
  \end{minipage}
  \begin{minipage}{0.48\textwidth}
    \includegraphics[width=\textwidth]{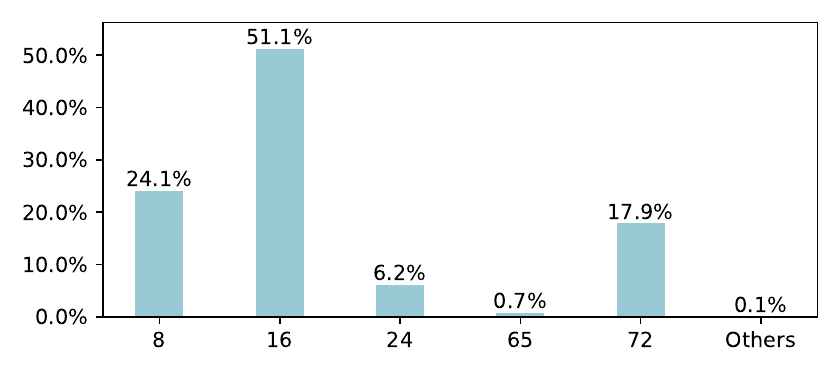}
    \vspace{-0.2in}
    \caption*{(b) generated videos}
  \end{minipage}
  \caption{The frame count distribution differences.}
  \label{fig:videos-frame}
  \vspace{-0.2in}
\end{figure}

\noindent\textbf{Frame Count Analysis.}
Figure \ref{fig:videos-frame} presents the distribution of frame count within our dataset. 
Due to the limitations of model capabilities and computational cost, the number of frames in real videos is significantly greater than in generated videos. 
The distribution of frame count in real videos is relatively even, with over 50\% of videos exceeding 120 frames. In contrast, generated videos are predominantly concentrated in the lower frame count range, with all videos having 72 frames or fewer.
The frame count of generated videos follows a distribution pattern based on fixed discrete values, while the frame count of real videos follows a various frames distribution.

\vspace{-0.1in}
\section{Method}
\vspace{-0.1in}

This section begins with the description of our network's design motivation (Section \ref{sec:motivation}). We then provide an overview of the dual-branch detector (DuB3D) architecture in Section\ref{sec:overall-archi}. Finally, we elaborate on the details of our network's implementation (Section ~\ref{sec:archi-details}).

\vspace{-0.1in}
\subsection{Motivation}
\label{sec:motivation}
\vspace{-0.1in}

General video detection models typically rely on evolving network structures, such as CNNs versus Transformers, and utilize techniques like skip connections. However, these modifications often overlook the specific characteristics of downstream tasks as they are designed for general purposes. In fake video detection tasks, our goal is to identify more discernible factors with tangible meanings and integrate them into the network architecture.

After conducting empirical experiments, we observed distinct differences in motion between real and fake videos. This is likely due to the generator's superior ability in appearance modeling compared to motion modeling. For instance, while the generator can synthesize a sequence of a human walking, it often struggles to accurately capture the pace, leading to the synthesis of abnormal poses.

Therefore, we propose a dual-branch structure consisting of an effective network for extracting features from the raw content of videos, alongside a motion-guided branch dedicated to capturing distinct motion characteristics. These two features represent the video's realism from different perspectives and can be fused in the deep feature space through neural transformation. We will describe this structure in detail in the following sections.

\vspace{-0.1in}
\subsection{Overall Architecture}
\label{sec:overall-archi}
\vspace{-0.1in}

As previously mentioned, videos offer not only extensive appearance information similar to images but also a wealth of temporal information capturing the movement of objects across frames, which is essential for video classification. Our approach integrates these two information sources through the Dual-Branch 3D Transformer we designed.

\paragraph{Dual-Branch 3D Transformer.}

To effectively capture both motion information and original spatial-temporal data, we propose the Dual-Branch 3D Transformer (DuB3D) architecture for classifying AI-generated videos.

\begin{figure}
  \centering
  \includegraphics[scale=0.42]{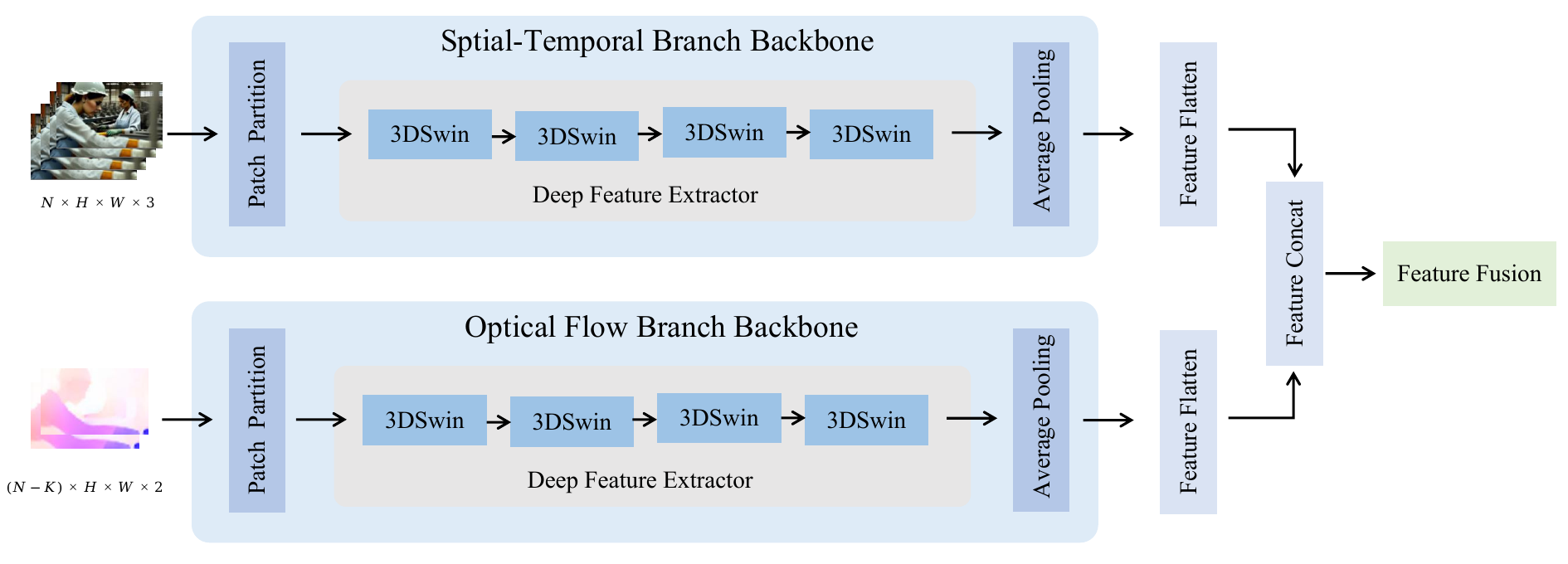}
  \caption{Overview of DuB3D architecture: 
  the upper branch represents the appearance modeling component, extracting spatial-temporal features from raw video content, while the bottom branch denotes the motion modeling component, acquiring motion features from optical flow. The upper branch processes $N$ frames, with an interval of $K$ frames used to compute optical flow for the input of the lower branch. In the network, ``3DSwin" refers to the Video Swin Transformer Stage~\cite{liu2021video}.
  }
  \label{fig:structure-1}
  \vspace{-0.2in}
\end{figure}

As depicted in Figure \ref{fig:structure-1}, this architecture comprises two branches: one for processing raw spatial-temporal data ($F_v$) and the other for optical flow ($F_o$). These branches are fused using a feature fusion block ($F_f$) at the end to facilitate classification. Our design utilizes the Video Swin Transformer (VideoSwin) as the backbone, tailored specifically for video recognition to leverage inherent spatiotemporal locality inductive bias.

\vspace{-0.1in}
\subsection{Arichitecture Details}
\label{sec:archi-details}

\paragraph{Sptial-Temporal Branch $F_v$.} 

This branch operates on original spatial-temporal frames to identify anomalies, such as wrongly generated artifacts in human bodies. Through this branch, the model effectively captures essential original spatial-temporal information for video comprehension. In practical applications, video clips contain multiple frames, and applying global attention across all frames can significantly increase computational costs. The Video Swin Transformer~\cite{liu2021video} introduces an inductive bias toward locality, improving the speed-accuracy trade-off, making it suitable as our backbone.

Specifically, we directly feed the original continuous frames into $F_v$ and obtain the output $f_v$ as
\begin{equation}
f_v=F_v(V),    
\end{equation}
where $V$ denotes the input video.

\paragraph{Optical Flow Branch $F_o$.} 
Learning-based optical flow estimation methods are essential for capturing video motion information. GMFlow~\cite{xu2022gmflow} is a popular and effective optical flow estimation model that reformulates optical flow as a global matching. We utilize GMFlow as the optical flow extractor in our architecture.
We define the original video $V$ has $N$ frames, and extract optical flow for each frames with a interval of $K$, as
\begin{equation}
W_i=M(V_i, V_{i+K}),
\end{equation}
where $V_i$ denotes the $i$-th frame in the video, $W_i$ is the estimated flow, and $M$ stands the optical flow estimation network like the GMFlow model. 
Then the optical flow's features can be obtained by forwarding the branch $F_o$ as 
\begin{equation}
    f_o=F_o(W_{i=1,...,N-K}),    
\end{equation}
where $f_o$ is the feature for the optical flow of these $N-K$ frames.

\paragraph{Final Fusion.}

Following the dual-branch setup, we construct a feature fusion block named ``Final Fusion`` comprised of a Multilayer Perceptron (MLP) to obtain the fused feature, as
\begin{equation}
    f=F_f(f_v \oplus f_o),
\end{equation}
where $\oplus$ is the concatenation operation and $F_f$ has the structure of a MLP with 3 layers.
The loss is computed between the classification feature $f$ and the ground truth using cross-entropy loss. Despite the simple structure of DuB3D, it underscores the significance of motion modeling in fake video detection and demonstrates excellent generalization ability across various types of real and generated videos, as verified in the subsequent experimental sections.

\vspace{-0.1in}
\section{Experiment}
\vspace{-0.1in}

\subsection{Evaluation Details}
\label{sec:eva-details}
\vspace{-0.1in}
\paragraph{Settings.}

All experiments were conducted using our GenVidDet dataset. The training dataset is a compilation of InternVid, HD-VG-130M, and VidProM, totaling 2.09 million clips (excluding Text2Video-Zero due to its limited frames). The validation dataset consists of two types: an in-domain test dataset, which includes Pika, ModelScope, and VideoCraft2, containing 173K clips specifically designed to measure the model's accuracy on generative models encountered during training, and an out-of-domain test dataset, which includes Open-Sora, Open-Sora-Plan, DynamiCrafter, and StreamingT2V, comprising 183K clips to evaluate the model's performance on generative models unseen during training.

\vspace{-0.15in}
\paragraph{Implementation Details.}

We begin by resizing the original video clips to 224 pixels along the shorter edge and cropping a central square with dimensions of 
$224 \times 224$ pixels. This resizing and cropping process ensures uniform input dimensions for subsequent analysis. To enhance dataset diversity and promote model generalization, we also apply random horizontal flips to the video frames. Prior to feeding the data into the network, we normalize it according to the prevailing ImageNet standards. For all experiments, we set the learning rate to 1.0e-4 with a scheduler decay of 0.925 after each 10\% steps in one epoch. Utilizing a batch size of 20, a weight decay of 5.0e-4 (1.0e-2 for the skip connection block), and a dropout rate of 0.25. We take the Swin-T~\cite{liu2021video} version as the backbones of both branches and train all weights from scratch. Experiments are performed on 3 machines with 6 A40 GPUs totally.

\vspace{-0.15in}
\paragraph{Metrics.}

We have selected Accuracy and F1-score as the evaluation metrics for our experiments. These metrics were chosen to emphasize both precision and recall, ensuring a comprehensive analysis of model performance. To ensure fair results, we predict the results of evaluation dataset firstly, then balance the real and generated samples to calculate evaluations for 10 times.

\subsection{Comparison}

In order to explore the optimal dual-branch architecture, we conduct experiments with four types of architectures, and results are shown in Table \ref{tab:exp-overall}.

\begin{table}
  \caption{Overall performance comparisons. ``Swapping feature'' and ``skip connection'' are described in Appendix \ref{sec:appendex_swapfeature} and \ref{sec:appendex_skipconnect}, and ``final fusion'' is described in the Section \ref{sec:archi-details}'s ``Final Fusion'' part. In methods of ``DuB3D-FF'' architecture, FI$i$ stands for ``with a frame interval of $i$''.}
  \centering
  \label{tab:exp-overall}
    
    \resizebox{1.0\linewidth}{!}{
  \begin{tabular}{p{5.5cm}<{\centering}p{5.5cm}<{\centering}p{1.5cm}<{\centering}p{1.5cm}<{\centering}p{1.5cm}<{\centering}p{1.5cm}<{\centering}}
    \toprule
    \multirow{2}{*}{Architecture}  & \multirow{2}{*}{Method} & \multicolumn{2}{c}{In-Domain} &\multicolumn{2}{c}{Out-of-Domain}   \\
    & & Accuracy & F1 & Accuracy & F1 \\
    \midrule
    \multirow{2}{*}{Single Branch} & Spatial-Temporal Branch & 0.9106 & 0.9118  & 0.6043  & 0.6858  \\
     & Union Branch & 0.9406   & 0.9414   & 0.6367   & 0.7074   \\
     \midrule
    \multirow{3}{*}{DuB3D with Swapping Feature (DuB3D-SF)} & DuB3D-SF (Bidirectional) & 0.9438  & 0.9444   & 0.7382   & 0.7675   \\
    & DuB3D-SF (Optical Flow)  & 0.9432  & 0.9435   & 0.6876   & 0.7345   \\
    & DuB3D-SF (Sptial-Temporal)  & 0.9509  & 0.9511   & 0.7057   & 0.7460   \\
    \midrule
    \multirow{2}{*}{DuB3D with Skip Connection (DuB3D-SC)} & DuB3D-SC (L2) & 0.9617   & 0.9618    & 0.7528    & 0.7784    \\
     & DuB3D-SC (L123) & 0.9654   & 0.9655   & 0.7709    & 0.7921    \\

    \midrule
    \multirow{3}{*}{DuB3D with Final Fusion (DuB3D-FF)} &DuB3D-FF (FI1) & \textbf{0.9677}  & \textbf{0.9679}   & 0.7521 & 0.7744 \\
    &DuB3D-FF (FI4) & 0.9523   & 0.9524  & 0.7271  & 0.7562 \\
    &DuB3D-FF (FI8) & 0.9570  & 0.9570   & \textbf{0.7919} & \textbf{0.7977} \\
    \bottomrule
    \end{tabular}
    }
    \vspace{-0.1in}
\end{table}


\vspace{-0.15in}
\paragraph{Motion Enhances Generated Videos Detection.} 

Table \ref{tab:exp-overall} presents the efficiency of optical flow in both single-branch and dual-branch architectures. In the Single Branch experiments group, the \textit{Spatial-Temporal Branch} is described in Section \ref{sec:archi-details}, while the \textit{Union Branch} concatenates spatial-temporal and optical flow data as input in one branch. 
Comparing the \textit{Spatial-Temporal Branch} method with the \textit{Union Branch} method highlights the efficiency of optical flow integration in the DAFV task. Additionally, comparing the \textit{Spatial-Temporal Branch} method with other methods of DuB3D, all results demonstrate that integrating optical flow improves performance compared to vanilla spatial-temporal 3D transformer.

\vspace{-0.15in}
\paragraph{Dual-Branch More Efficient Than Single Branch.} 

Comparing the \textit{Single Branch} architecture with the three DuB3D architectures featuring different feature fusion methods, each DuB3D architecture significantly outperforms the \textit{Single Branch} architecture in video detection. Additionally, comparing the \textit{Union Branch} method with the methods of DuB3D, all of which involve feeding spatial-temporal and motion data, we find that splitting the spatial-temporal and motion data into two branches with either backbone allows the model to learn deep features more effectively. 
This suggests that spatial-temporal data and motion data have distinct distinguishing features, and feeding them into unique backbones improves the extraction of deep features. Furthermore, in out-of-domain results, the dual-branch architecture exhibits better efficiency improvement than in-domain detection, indicating that dual-branch architectures can especially enhance the generalization of detectors.

\vspace{-0.15in}
\paragraph{More Connection Not Always Better.} 
To explore the strength of connection between dual-branch architectures, we experiment with three types of connections between branches. In \textit{DuB3D with Swapping Feature}, we swap the output of the inner stage of 3D Transformer blocks between $F_v$ and $F_o$, feeding the swapped features into the next stage of the other backbone. In \textit{DuB3D with Skip Connection}, we implement skip connections from the inner stages of $F_v$ and $F_o$ to the final feature fusion layer. Wondrously, the native \textit{DuB3D} with the ``Final Fusion" stage alone demonstrates the best performance. From these comparisons, we observe that extracting deep features independently from each branch is sufficient; skip connections and feature swapping do not significantly improve performance. Therefore, considering the cost, there is no need to perform feature fusion in the inner layers of the two branches.
The setting details of ``with Swapping Feature" and ``with Skip Connection" can be viewed in Appendix \ref{sec:appendex_swapfeature} and \ref{sec:appendex_skipconnect}.

\subsection{Ablation Study}


\vspace{-0.1in}
\paragraph{More Frame Count Improves Performance.}

In addition to conducting experiments with a 16-frame count as shown in Table \ref{tab:exp-overall}, we also explored the influence of frame count by conducting experiments with a 12-frame count. The results in Table \ref{tab:frame_count} indicate that a higher frame count can improve detection performance in both in-domain and out-of-domain evaluations. Furthermore, for out-of-domain video detection, a higher frame count can significantly enhance performance.

\begin{table}
  \caption{Performance with different frame counts.}
   \label{tab:frame_count}
  \centering
  \resizebox{1.0\linewidth}{!}{
  \begin{tabular}{p{5.5cm}<{\centering}p{2.5cm}<{\centering}p{2.5cm}<{\centering}p{2.5cm}<{\centering}p{2.5cm}<{\centering}}
    \toprule
    \multirow{2}{*}{Method} & \multicolumn{2}{c}{In-Domain} &\multicolumn{2}{c}{Out-of-Domain}   \\
    & Accuracy & F1-score & Accuracy & F1-score \\
    \midrule
    DuB3D-FF (FI8) with 12 Frames& 0.9520  & 0.9522  & 0.7185  & 0.7483  \\
    DuB3D-FF (FI8) with 16 Frames& \textbf{0.9570} & \textbf{0.9570}   & \textbf{0.7919} & \textbf{0.7977} \\
    \bottomrule
  \end{tabular}
  }

    \caption{Performance with different frame rates.}
  \label{tab:frame_rate}
  \centering
  \resizebox{1.0\linewidth}{!}{
  \begin{tabular}{p{5.5cm}<{\centering}p{2.5cm}<{\centering}p{2.5cm}<{\centering}p{2.5cm}<{\centering}p{2.5cm}<{\centering}}
    \toprule
    \multirow{2}{*}{Method} & \multicolumn{2}{c}{In-Domain} &\multicolumn{2}{c}{Out-of-Domain}   \\
    & Accuracy & F1-score & Accuracy & F1-score \\
    \midrule
    DuB3D-FF (FI8) with original fps& \textbf{0.9643}  & \textbf{0.9644}  & 0.7200  & 0.7626  \\
    DuB3D-FF (FI8) with 8 fps& 0.9570  & 0.9570   & \textbf{0.7919} & \textbf{0.7977} \\
    \bottomrule
  \end{tabular}}
  \vspace{-0.1in}
\end{table}

\vspace{-0.15in}
\paragraph{Scaling Frame Rate Benefits Out-of-Domain Detection.} 

Since different models generate videos with varying frame rates as shown in Table \ref{tab:Generating-Videos-Curation}, whereas real videos are typically produced at frame rates ranging from 24 to 30 fps, we experimented with both the original frame rate and scaling the frame rate to 8 fps to explore the impact of frame rate. The results in Table \ref{tab:frame_rate} demonstrate that scaling the frame rate to a uniform value can improve performance on out-of-domain video detection.

\vspace{-0.15in}
\paragraph{Impact of Frame Interval in Optical Flow.} 

In the \textit{DuB3D with Final Fusion} architecture of Table \ref{tab:exp-overall}, we conducted experiments with frame intervals of $1/4/8$ to extract optical flow. The overall results in Table \ref{tab:exp-overall} and the detailed results in Table \ref{tab:all_models} both demonstrate that a larger frame interval can improve generalization, while a smaller frame interval is better suited for fitting the in-domain dataset. We recommend using a sufficient frame interval for optical flow estimation to achieve better generalization in realistic applications.

\begin{table}
  \caption{Performance on different generated models.}
  \label{tab:all_models}
  \centering

    \resizebox{1.0\linewidth}{!}{
      \begin{tabular}{p{2.5cm}<{\centering}p{2.5cm}<{\centering}p{2.5cm}<{\centering}p{2.5cm}<{\centering}p{2cm}<{\centering}p{2.5cm}<{\centering}p{2cm}<{\centering}p{2cm}<{\centering}}
    \toprule
    \multirow{2}{*}{Method} & \multicolumn{3}{c}{In-Domain (F1-score/Accuracy)} &\multicolumn{4}{c}{Out-of-Domain (F1-score/Accuracy)}   \\
    & Pika & ModelScope & VideoCraft2 & Open-Sora & Open-Sora-Plan & DynamiCrafter & StreamingT2V \\
    \midrule
    DuB3D-FF (FI1) & \textbf{0.9562}/\textbf{0.9560}  & \textbf{0.9809}/\textbf{0.9808}   & 0.9738/0.9739 & 0.7254/\textbf{0.6659} & 0.7923/0.7697 & 0.8348/0.8284  & 0.8793/0.8749 \\
    DuB3D-FF (FI4) & 0.9371/0.9367 & 0.9660/0.9657  & 0.9602/0.9601 & 0.7037/0.6236 & 0.7670/0.7380 & 0.8096/0.7991 & \textbf{0.9227}/\textbf{0.9228} \\
    DuB3D-FF (FI8) & 0.9352/0.9348   & 0.9745/0.9745  & \textbf{0.9769}/\textbf{0.9770} & \textbf{0.7361}/0.6611 & \textbf{0.7940}/\textbf{0.7757}  & \textbf{0.8896}/\textbf{0.8882}  & 0.8313/0.8188  \\
    \bottomrule
  \end{tabular}
    }

\caption{Performance with different dataset scales. All the experiments are conducted on DuB3D-FF (FI8), and evaluated on Out-Of-Domain dataset.}
  \label{tab:sample_count}
  \centering
    \resizebox{1.0\linewidth}{!}{
  \begin{tabular}{p{3cm}<{\centering}p{2.5cm}<{\centering}p{2cm}<{\centering}p{2cm}<{\centering}p{2.5cm}<{\centering}p{2.5cm}<{\centering}p{2.5cm}<{\centering}p{2.5cm}<{\centering}}
    \toprule
    Method &Sample Rate & Data Size & Epoch  & Accuracy &   F1-score \\
    \midrule
    DuB3D-FF (FI8) & 10\% & 209K & 10 &  0.7202  & 0.7565  \\
    DuB3D-FF (FI8) & 100\% & 2.09M & 1&  \textbf{0.7919} & \textbf{0.7977} \\
    \bottomrule
  \end{tabular}}
  \vspace{-0.1in}
\end{table}

\vspace{-0.15in}
\paragraph{Large-Scale Dataset Improve Generalization Performance Significantly.}

We conducted an experiment to verify the assumption that a larger-scale dataset can improve performance. We randomly sampled 10\% of the samples from our training set resulting in 209K samples which is enough for fake video detection tasks commonly(e.g. DFDC~\cite{dolhansky2020deepfake} dataset has 128K clips), and ran 10 epochs using the formal experiment's settings for a fair comparison. Table \ref{tab:sample_count} demonstrates that a larger dataset size significantly improves out-of-domain performance.

\section{Conclusion and Future Work}
\label{sec:conclusion}

In this paper, we introduce GenVidDet, the first dataset comprising over \textbf{2.66 million} instances of both real and generated videos, varying in categories, frames per second, resolutions, and lengths. Additionally, we present DuB3D, an innovative and effective method for separating real and generated videos. Our extensive experiments on GenVidDet demonstrate the effectiveness of our proposed method DuB3D. Despite advancements, there are still improvement rooms. E.g., while we have investigated the 3D transformer approach, further exploration of other efficient techniques is warranted.

\bibliographystyle{plainnat}
\bibliography{dub3d}

\clearpage
\appendix


\section{The Answer of Figure \ref{fig:game}}
\label{sec:appendex_game}

Table \ref{tab:game-answer} presents the answers corresponding to Figure \ref{fig:game}. This example demonstrates that generated videos have become very difficult to distinguish from real videos.

\begin{table}
    \tiny
  \caption{The answer of figure \ref{fig:game}}
  \label{tab:game-answer}
  \centering
  \resizebox{1.0\linewidth}{!}{
  \begin{tabular}{p{0.5cm}<{\centering}p{1.2cm}<{\centering}p{1.2cm}<{\centering}p{5cm}<{\centering}}
    \toprule
    Choice & Answer & Source & Details \\
    \midrule
    A & Generated & Pika & \textbf{Prompt}: Slow motion, wild Fire and smoke movement in the background -camera zoom in, -neg morphing, erratic fluctuation in motion, noisy, bad quality, distorted, poorly drawn, blurry, grainy, low resolution, overesaturated, lack of detailed, inconsistent lighting \\
    \midrule
    B & Real & Youtube & \textbf{URL}: https://www.youtube.com/watch?v=-gv8dh-GfYE \\
    \midrule
    C & Real & Youtube & \textbf{URL}: https://www.youtube.com/watch?v=3tjxe-EBKa8 \\
    \midrule
    D & Generated & StreamingT2V & \textbf{Prompt}: person pouring lemonade into a plastic container of water \\
    \midrule
    E & Generated & Open-Sora-Plan & \textbf{Prompt}: pomeranian puppy in the bathroom looking at the shower mat \\
    \bottomrule
  \end{tabular}
  }
\end{table}

\section{DuB3D with Swapping Feature}
\label{sec:appendex_swapfeature}

\begin{figure}
  \centering
  \includegraphics[scale=0.44]{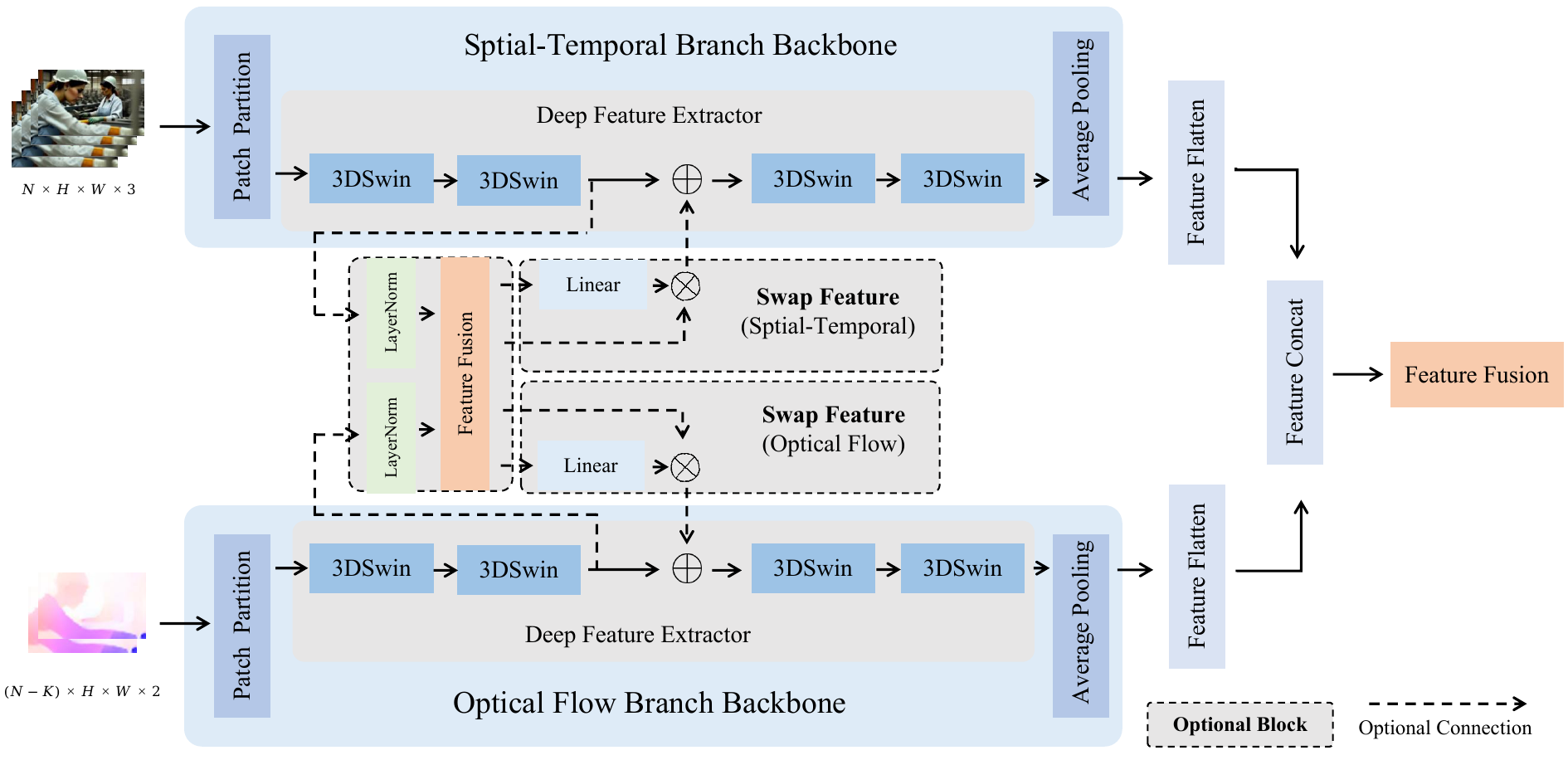}
  \vspace{-0.15in}
  \caption{Architecture of DuB3D with Swapping Feature (Bidirectional)}
  \label{fig:swap-feature}
  \vspace{-0.1in}
\end{figure}

To explore the strength of feature fusion in inner layers between dual-branch, we design a \textbf{Swap Feature Block} between two branch backbones. As is shown in Figure \ref{fig:swap-feature}, feeding with the output of the $2$-th \textit{3DSwin} we fuse the features from each other. Then through a \textit{Linear Layer} to control the fusion weights, we swap the fusion features into the other backbone.
According to the \textit{Swap Feature Block} selection, we design three experiments as below:
\begin{itemize}
    \item[$\bullet$] \textbf{DuB3D-SF (Bidirectional)}: DuB3D with Swapping Feature (Bidirectional) which is shown as Figure \ref{fig:swap-feature} applying \textit{Swap Feature (Spatial-Temporal)} and \textit{Swap Feature (Optical Flow)}.
    
    \item[$\bullet$] \textbf{DuB3D-SF (Spatial-Temporal)}: DuB3D with Swapping Feature (Spatial-Temporal) which only utilizes \textit{Swap Feature (Spatial-Temporary) Block} in the Figure \ref{fig:swap-feature}.

    \item[$\bullet$] \textbf{DuB3D-SF (Optical Flow)}: DuB3D with Swapping Feature (Optical Flow) which only utilizes \textit{Swap Feature (Optical Flow) Block} in the Figure \ref{fig:swap-feature}.
    
\end{itemize}

All experiments are conducted with frame interval of 8, and the results for this architecture are shown in Table \ref{tab:exp-overall}, we observe that swapping features between branches can not significantly improve performance.  

\section{DuB3D with Skip Connection}
\label{sec:appendex_skipconnect}

\begin{figure}[htbp]
  \centering
  \includegraphics[scale=0.44]{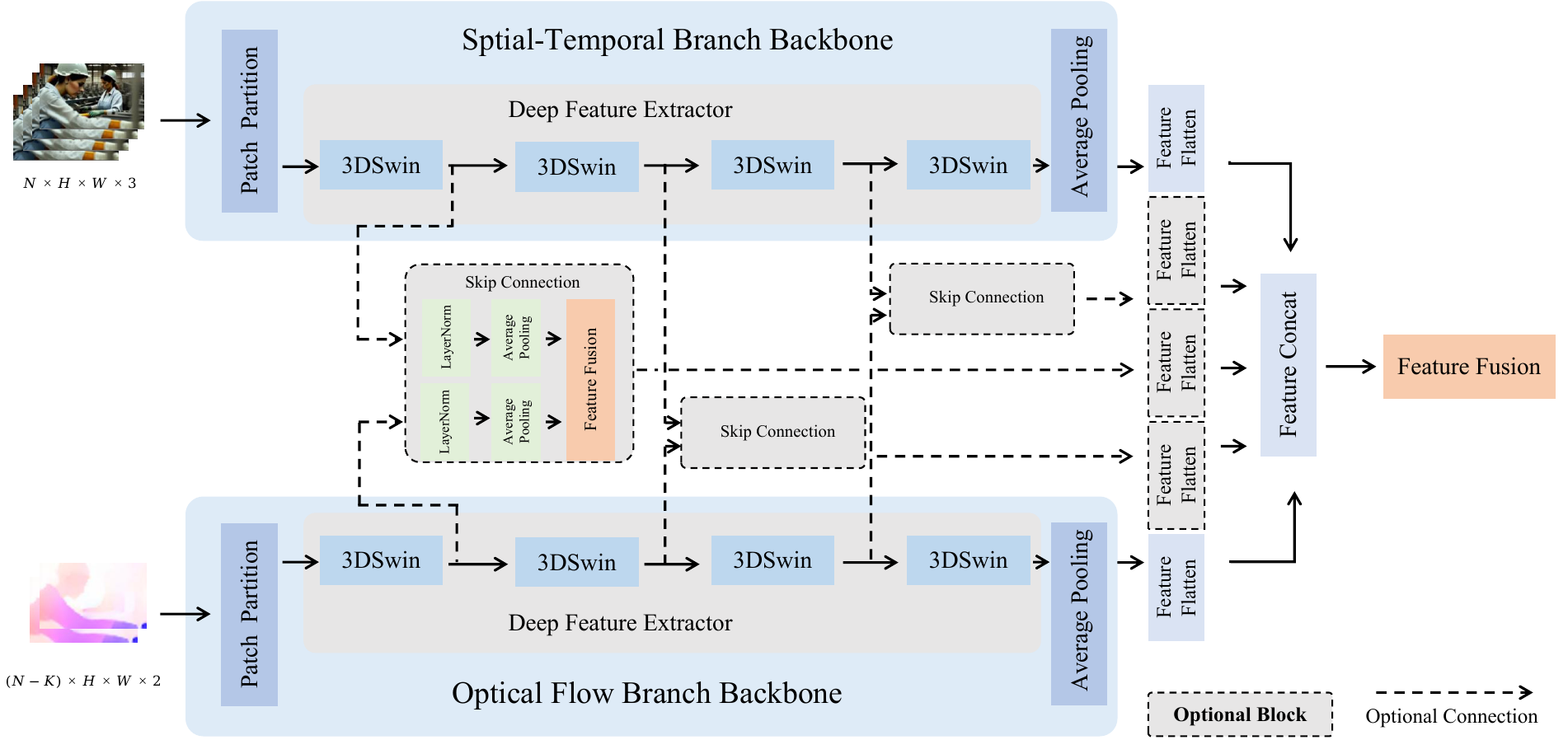}
  \vspace{-0.15in}
  \caption{Architecture of DuB3D with Skip Connection}
  \label{fig:skip-connection}
  \vspace{-0.1in}
\end{figure}

To explore the impact of skip connections directly to the final feature fusion between two branches, we design \textit{DuB3D with Skip Connection} architecture shown in Figure \ref{fig:skip-connection}. The outputs of \textit{3DSwin} in both branches are applied with \textit{LayerNorm} and \textit{Average Pooling}, then fused by \textit{Feature Fusion} described as Section \ref{sec:archi-details}. Based on the optional \textit{Skip Connection Block}, we design the following experiments:
\begin{itemize}
    \item[$\bullet$] \textbf{DuB3D-SC (L2)}: which only add \textit{Skip Connection Block} following the output of the $2$-nd \textit{3DSwin}. 
    \item[$\bullet$] \textbf{DuB3D-SC (L123)}: This setup adds a \textit{Skip Connection Block} after the output of each of the $1$-st, $2$-nd, and $3$-rd \textit{3DSwin} layers. 
\end{itemize}
All experiments are conducted with frame interval of 8. As is shown in Table \ref{tab:all_models}, the results reveals that skip connection can gain good result, but can not improve performance significantly.

\end{document}